%% file: main.tex
\documentclass[runningheads]{llncs}
\usepackage[T1]{fontenc}
\usepackage{graphicx}
\usepackage{amsmath}
\usepackage{subcaption}
\usepackage{tikz, pgfplots}
\pgfplotsset{compat=1.18}
\usetikzlibrary{fit}
\usetikzlibrary{spy}
\usepackage{booktabs}
\usepackage{multirow}
\usepackage{color}

\begin{document}
\title{Transparency Distortion Robustness for SOTA Image Segmentation Tasks}
%
%
\author{Volker Knauthe\inst{1}\orcidID{0000-0001-6993-5099} \and
 Arne Rak\inst{1}\orcidID{0000-0001-6385-3455} \and
 Tristan Wirth\inst{1}\orcidID{0000-0002-2445-9081} \and
 Thomas P\"ollabauer\inst{2,1}\orcidID{0000-0003-0075-1181} \and
 Simon Metzler\inst{1} \and
 Arjan Kuijper\inst{2,1}\orcidID{0000-0002-6413-0061} \and
 Dieter\,W. Fellner\inst{1,2,3}\orcidID{0000-0001-7756-0901}}

 \authorrunning{V. Knauthe et al.}
 
%
 \institute{Technical University of Darmstadt, Darmstadt, Germany \and
 Fraunhofer Institute for Computer Graphics Research IGD, Darmstadt, Germany \and
 CGV Institute, Graz University of Technology, Graz, Austria}
\maketitle              
\begin{abstract}
Semantic Image Segmentation facilitates a multitude of real-world applications ranging from autonomous driving over industrial process supervision to vision aids for human beings.
These models are usually trained in a supervised fashion using example inputs.
Distribution Shifts between these examples and the inputs in operation may cause erroneous segmentations.
The robustness of semantic segmentation models against distribution shifts caused by differing camera or lighting setups, lens distortions, adversarial inputs and image corruptions has been topic of recent research.
However, robustness against spatially varying radial distortion effects that can be caused by uneven glass structures (e.g. windows) or the chaotic refraction in heated air has not been addressed by the research community yet.
We propose a method to synthetically augment existing datasets with spatially varying distortions.
Our experiments show, that these distortion effects degrade the performance of state-of-the-art segmentation models.
Pretraining and enlarged model capacities proof to be suitable strategies for mitigating performance degradation to some degree, while fine-tuning on distorted images only leads to marginal performance improvements.

\end{abstract}
\input{commands}

\input{sec/intro}

\input{sec/related_work}

\input{sec/experiments}

\input{sec/results}

\input{sec/distortiontraining}

\input{sec/conclusion}

\begin{credits}
\end{credits}
%
%
%
\bibliographystyle{splncs04}
\bibliography{bibliography}
\end{document}

%% file: commands.tex
\definecolor{colorscheme0}{HTML}{009392}
\definecolor{colorscheme1}{HTML}{39b1b5}
\definecolor{colorscheme2}{HTML}{9ccb86}
\definecolor{colorscheme3}{HTML}{e9e29c}
\definecolor{colorscheme4}{HTML}{eeb479}
\definecolor{colorscheme5}{HTML}{e88471}

\newcommand{\evalcrop}[6]{
    \begin{tikzpicture} [spy using outlines={very thick, rectangle,yellow,magnification=2.5, connect spies}, inner sep=0pt, outer sep=0pt]
        \node[anchor=#6] (im) at (0, 0)
        {\includegraphics[width=\linewidth]{#1}};
        \spy[anchor=#6, draw, height=#5, width=#4] on (#2, #3) in node (zoom) at (0, 0);
    \end{tikzpicture}%
}

\newcommand{\evalcropcrop}[7]{
    \begin{tikzpicture} [spy using outlines={very thick, rectangle,yellow,magnification=2.5, connect spies}, inner sep=0pt, outer sep=0pt]
        \node[anchor=#6] (im) at (0, 0)
        {\includegraphics[width=\linewidth]{#1}};
        \node[anchor=south west] (im) at (-3.9, 0)
        {\includegraphics[width=.465\linewidth]{#7}};
        \spy[anchor=#6, draw, height=#5, width=#4] on (#2, #3) in node (zoom) at (0, 0);
    \end{tikzpicture}%
}

\pgfplotsset{
    mybarplot/.style={
        width=\linewidth,height=5cm,
        ybar=0pt,
        scale only axis,
        cycle list={colorscheme0, colorscheme1, colorscheme2, colorscheme4, colorscheme5},
        ytick style={/pgfplots/minor tick length=0pt },
        axis on top,
        every axis plot post/.append style={
            fill,
            draw=none,
        }
    },
            error bars/.cd, y dir=both, y explicit, error bar style={color=black}
}

%% file: sec/intro.tex
\section{Introduction}
\textit{Computer Vision Algorithms} have reached an impressive level of performance in various tasks, which leads to a rising number of real world applications.
Especially \textit{Semantic Image Segmentation} is a high level technique, that benefits a large area of tasks like autonomous agent world perception, industrial process supervision and vision aids for humans with strong impairments.
While there is a lot of work regarding the general robustness of computer vision models against environmental influences \cite{hendrycks2019benchmarking25,hendrycks2021natural} and global distortions \cite{ye2020universal,deng2019restricted,xu2019semantic,zhang2022bending,zheng2023look}, transparency induced spatially varying distortions are yet underrepresented.
Common real-world occurrences are glass structures in cities and public buildings like hospitals, as well as the ever present air itself.
This becomes apparent on hot summer days, in chemical labors or production lines that operate on high temperatures.
While not every transparency necessarily leads to distortion, heat, bent glass, worn glass and not optimised glass sooner or later introduce irregular patterns.
Some transparent objects, like lenses or art/interior, even distort by nature.\\
The main contributions of this paper consists of a method to synthetically generate locally distorted images (grid distortion model) and an evaluation of how state-of-the-art image segmentation models perform these distorted images.
The distortion operation simulates common transparency behaviour and can be easily applied on existing datasets. 
The evaluation inspects different variations of Swin Transformer \cite{liu2021swin} and includes a comparison with DINOv2 \cite{oquab2023dinov2}. 
We show, that local distortions from transparent structures can noticeably inhibit semantic segmentation capabilities and that fine-tuning and retraining wont solve this challenge. 
Furthermore, different model properties, like network and training data size only increase the model's robustness against certaina
distortion types. 
However, an effective way to increase the robustness is via more extensive pre-training on bigger datasets like that of DINOv2.


%% file: sec/related_work.tex
\section{Related Work}

In this chapter we discuss the recent advancements and the state-of-the-art in semantic image segmentation (section \ref{sec:rel:semantic}) and machine learning robustness (section \ref{sec:rel:robustness}) research. We further give an overview over the sparse literature situation regarding robustness against distortions (section \ref{sec:rel:distortion}). To the best of our knowledge, local distortion robustness has not been addressed yet.

\subsection{Semantic Segmentation}
\label{sec:rel:semantic}

Semantic Segmentation describes the task of assigning separate class labels to each pixel of an input 2D image \cite{ulku2022survey}.
This constitutes an important sub task in a multitude of real-world applications, such as autonomous driving, robot-assisted surgery and the creation of robots that directly interact with human beings.
With the emergence of Convolutional Neural Networks (CNNs) semantic segmentation has made a leap in the resulting classification quality \cite{shelhamer2017fully}.
Encoder-Decoder architectures \cite{ronneberger2015u,badrinarayanan2017segnet,ulku2020comparison,iglovikov2018ternausnetv2} have also shown significant performance improvements in the context of semantic image segmentation.
These strategies have been further refined by spatial pyramid pooling \cite{he2015spatial,li2018pyramid}, that incorporates features from different levels within the feature extractor to enhance predictions.
The introduction of the attention mechanism, e.g., Vision Transformers \cite{dosovitskiy2020image} (ViT), further increases the segmentation quality.
These methods calculate the influence of image patches on one another, bringing global information into the underlying CNN.
Swin Transformers \cite{liu2021swin} further improve that strategy by incorporating transformers hierarchically, i.e., not only on the global image but also further subdividing patches into sub patches.
Most recently, foundation models, such as the Segment Anything Model \cite{kirillov2023segment} (SAM) and DINOv2 \cite{oquab2023dinov2} have emerged. These models utilize vast training data sets to train effective feature extractors, that enhance subsequent neural networks' performance on a multitude of vision tasks including semantic segmentation.
DINOv2 is a set of pretrained visual models trained with different Vision Transformer (ViT) architectures \cite{dosovitskiy2020image}, that can be used to improve the performance on a multitude of vision tasks.
It is based on a combination of DINO \cite{caron2021emerging} and iBOT \cite{zhou2021ibot}, which introduces masked image modeling (MLM), with additional modifications.
The SAM \cite{kirillov2023segment} is specifically trained for the task of image segmentation.
The model is promptable in order to transfer zero-shot examples to new image distributions and tasks.

\subsection{Robustness}
\label{sec:rel:robustness}


Although the robustness of machine learning models is extensively discussed, there is no coherent definition of the term robustness \cite{drenkow2021systematic,liu2023comprehensive}.
Most robustness literature is concerned with the mitigation of adversarial inputs \cite{drenkow2021systematic}, that are subtly altered by an adversary in a way that the modifications are imperceptible to humans, yet they cause a neural network to alter its prediction \cite{szegedy2014intriguing}.
Robustness also often refers to situations where distribution shifts \cite{quinonero2008dataset} occur between training examples and operation.
Notable examples for these distribution shifts are changes in environmental setting that can be caused by a multitude of variables, such as time, camera location, etc. \cite{hendrycks2021natural,recht2019imagenet}.
The challenge discussed in this paper, where distortions caused by transparent objects influence the appearance of the input images, is part of this sub class of robustness challenges.
Hendrycks and Dietterich \cite{hendrycks2019benchmarking25} introduce the terms \textit{corruption robustness}, which refers to a models ability the give correct predictions for images that contain artifacts or distortion, and \textit{perturbation robustness}, which refers to the stability of a model's prediction for minor changes in the input image.
Freiesleben and Grote \cite{freiesleben2023beyond} summarize the different notions of robustness in the machine learning literature by formally defining it as \textit{the relative stability of a robustness target with respect to specific interventions on a modifier}.\\
Artificial corruptions such as blur, noise and brightness changes significantly decrease the accuracy of neural networks on image classification tasks \cite{hendrycks2021many24,hendrycks2021natural,mu2019mnist39}.
The application of data augmentation mitigates these effects by inserting artificially corrupted images into the training set \cite{perez2017effectiveness42}.
However, these mitigation strategies do not generalize well from one corruption type to another \cite{dodge2017quality14,geirhos2018generalisation20,mu2019mnist39}.
Furthermore, Taori et al. \cite{taori2020measuring53} suggest that synthetic image corruptions do not always transfer well to the corruptions occurring in the real world.\\
Some strategies try to increase the robustness of learned features.
Geirhos et al. \cite{geirhos2018imagenet19} consider style transfer making the resulting models less susceptible to minor variations in the texture information, and instead more reliant on shape information.
RobustNet \cite{choi2021robustnet7} uses feature correlations to disentangle style information and relevant task content.
Zheng et al. \cite{zheng2016improving63} propose a stability loss, that is targeted to mitigate adversarial attacks by adding small perturbations to input images and enforcing similar outputs for these examples.\\
Some contributions also discuss the impact of the chosen model architecture for model robustness.
Arnab at al. \cite{arnab2018robustness2} show that ResNets' \cite{he2016deep23} robustness exceeds the robustness of previous architectures like VGG \cite{simonyan2014very51} on the task of semantic segmentation.
Hendryck et al. \cite{hendrycks2021natural} confirm this finding regarding image classification by showing that both the ResNet architecture \cite{xie2017aggregated60} and self-attention \cite{hu2018squeeze28,touvron2021training54} improve model robustness.\\
Increasing the amount of training data has also shown to be beneficial for model robustness \cite{geirhos2021partial18}.
However, the required amount of additional data is quite high \cite{hendrycks2021natural}, making this approach costly.
The foundation model DINOv2 \cite{oquab2023dinov2} shows a significant performance increase on standard robustness benchmarks for image classification such as ImageNet-A \cite{hendrycks2021natural} and ImageNet-C \cite{hendrycks2019benchmarking25}.
The underlying vast pool of training data used for the Segment Anything Model\cite{kirillov2023segment} (SAM) is also shown to have an increased prediction performance both on images from the training distribution as well as out of distribution examples.\\
The conducted research on semantic segmentation is sparse comparatively sparse.
Arnab et al. \cite{arnab2018robustness2} review the robustness of semantic segmentation models towards adversarial attacks.
Kamann et al. \cite{kamann2020benchmarking} conducted a study in which they show that general model performance is a good predictor for robustness for segmentation tasks based on an ablation of DeepLabv3+ \cite{chen2018encoder}. Furthermore, they show that architectural choices can have significant influence.
However, since these works predate the introduction of transformer based architectures we consider them outdated to some extent.

\subsection{Distortion Robustness}
\label{sec:rel:distortion}

In this work, we examine the robustness of state-of-the-art techniques towards natural occurring spatially varying distortions.
These kinds of distortion can be caused by, e.g., uneven transparent structures or chaotic light refraction in heated air.
This further expands our understanding of deep learning model robustness for different kinds of natural distribution shifts.\\
Recent research on distortion robustness of deep learning models regarding distortions mainly focuses on global distortions.
Ye et al. \cite{ye2020universal} show that training on globally distorted views enhances semantic segmentation performance on images captured by fish eye lenses.
Deng et al. \cite{deng2019restricted} introduce deformable convolutions to address that task.
Xu et al. \cite{xu2019semantic} show that artificially creating panoramic images as training data enhances performance of semantic segmentation on panoramic images.
Subsequent work shows that architectural enhancements such as the introduction of distortion-aware transformers \cite{zhang2022bending} and leveraging neighborhood information on pixel level \cite{zheng2023look} further improve segmentation results on that task.\\
The research on robustness against local distortions is limited to the mitigation of the effects of raindrops on windows and windshields.
Halimeh and Roser \cite{halimeh2009raindrop22} use explicit geometric modeling to address that task.
Eigen et al. \cite{eigen2013restoring17} solve this problem with the use of Convolutional Neural Networks.
Quan et al. \cite{quan2019deep44} enhance performance of the resulting images using attention models.
However, these publications do not explicitly address robustness of the models, but rather have the goal to reconstruct distortion free images.
Further work \cite{knauthe2023distortion} shows, that the detection of transparency induced distortions in images, are hard to detect for humans, posing a relevant problem in the real world.\\
To our knowledge, our contribution is the first work, that examines the robustness of semantic segmentation models on spatially varying distortions such as those imposed from uneven windows, windshields or heated air explicitly.

%% file: sec/experiments.tex
\section{Distortion Robustness Experiments}

In this work, we create a dataset of artificially distorted images with spatially varying distortion as real world capture is impractical, due to the complexity and time needed for their collection.
For this, the ADE20K dataset \cite{zhou2019semantic} is chosen as a basis, as it covers diverse indoor and outdoor scenes where transparent structures commonly exist. In section \ref{sec:dataset} our dataset generation process is covered, which involves a novel distortion model.
Furthermore, we identify suitable parameters for our distortion model through rigorous experiments on the ADE20K dataset.
We elaborate our model selection process in section \ref{sec:image-seg-method}.

\input{sec/figure_distortion_method}

\subsection{Dataset and Distortion Parameters} \label{sec:dataset}

While prior studies extensively investigate the effects of different lens distortions such as fisheye \cite{ye2020universal,deng2019restricted} and panoramic views \cite{xu2019semantic,zhang2022bending,zheng2023look}, there is very little understanding of robustness to spatially varying distortions caused by general transparent structures. 
To address this gap, we simulate an environment where a picture is taken through a transparent structure like an uneven window pane that distorts the image.
For this, we employ two distortion models that are based on a radial distortion model (see Eq. \ref{eq:distortion}) that is derived from Brown \cite{brown1971close} by omitting the tangential distortion in accordance with other distortion datasets \cite{liao2020drgan}.
Global radial distortion simulates one lens that covers the whole image area.
Grid radial distortion assumes spatially varying radial distortion on patches of an image that are arranged in a $10\times 10$ grid, where the patch size is dynamically adjusted to the image's resolution.
This method is based on the idea of heat shimmers and/or cameras in front of window panes, which can result in a variety of local distortions.
Each patch is distorted with different parameters, sampled from a uniform random distribution.
Between the patches, the pixel shift is interpolated between all adjacent distortions in 10px wide strips. 
The interpolation mitigates gaps and facilitates smooth transitions between patches so that all objects and lines that are connected in the undistorted image remain connected after the distortion. 
This results in a distortion effect that closely resembles that of an unevenly thick window pane or heated air.
Figure \ref{fig:grid_distortion} illustrates the effect of our grid distortion model on a uniform grid pattern and a real scene. 
\begin{align}
x_u = x_d (1 + K_1 r^2 + K_2 r^4) && y_u = y_d (1 + K_1 r^2 + K_2 r^4)
\label{eq:distortion}
\end{align}
In Eq. \ref{eq:distortion}, $x_d, y_d \in [-1, 1]$ and $x_u, y_u$ denote the distorted, normalized and undistorted image coordinates respectively, $r$ denotes the distance of the distorted image point to the distortion center and $K_n$ denote the radial distortion coefficients.
To determine realistic values for $K_1$ and $K_2$, several parameter configurations are evaluated on the ADE20K dataset. \\
We initially assume the parameter range to be $[-0.5, 0.5]$.
To further refine this range, we qualitatively evaluate various parameter configurations on images of the ADE20K dataset, an example of which is shown in Fig. \ref{fig:both_dists} (a).
Figure \ref{fig:both_dists} (b) and (c) illustrate the distortive effects of different values of $K_n$ for the global distortion model.
For $K_n > 0$, a barrel distortion effect is achieved.
At values beyond 0.4, significant image loss occurs due to excessive cropping, rendering higher distortion values impractical.
The pincushion distortion effect achieved with negative values $K_n$ exhibits more severe warping in the image.
Even with $K_2$ fixed to zero, substantial distortion can be observed for $K_1 < -0.2$, leading to artifacts where objects are stretched so far that they are hardly recognizable.
Considering realism, values beyond -0.2 for $K_1$ are deemed impractical, while $K_2$ is restricted to positive values for the global distortion model.
\\
Figure \ref{fig:both_dists} (d)-(f) illustrate the distortive effects for different values of $K_n$ for the grid distortion model.
As strips between the grid patches are interpolated, the severe warping at image boundaries observed for the global distortion model is reduced here.
Consequently, the previously established parameter range limitations are lifted.
Parameter values $|K_n| > 0.5$ introduce frequently occurring artifacts such as disconnected object contours (see Fig. \ref{fig:both_dists}f), which may occur naturally, but are considered out of scope for this work.
\\
With practical parameter ranges identified for both the global and grid distortion models, we define value ranges for $K_1$ and $K_2$ based on a hyperparameter $\sigma$ that we call distortion intensity. $\sigma$ is linearly correlated with the mean pixel shift, i.e., the mean Euclidean distance between shifted pixel positions and their origin location. In our experiments, we use $K_1 \in [-\frac{\sigma}{2},\sigma]$ and $K_2 
\in [0,\sigma]$ with $\sigma \in [0, 0.4]$ for global distortion and $K_1, K_2 \in [-\sigma,\sigma]$ with $\sigma \in [0, 0.5]$ for the grid distortion. Parameters $K_n$ are sampled from random uniform distributions adhering to their respective value ranges.

\input{sec/figure_global_distortion}

\subsection{Image Segmentation Method}
\label{sec:image-seg-method}

For our experiments, we employ Swin Transformer \cite{liu2021swin} based on its good performance on semantic segmentation tasks and due to the recent shift of computer vision backbone models towards transformers. 
Previous robustness research \cite{oquab2023dinov2,hendrycks2021natural,hendrycks2019benchmarking25} shows that model capacity and pretraining can have an impact on robustness.
Thus, both the tiny version Swin-T with 60 million parameters and the base version Swin-B with 121 million parameters are leveraged for evaluation. 
Both models are trained on the ADE20K dataset. 
For analyzing the influence of pretraining, we compare performance of models Swin-T and Swin-B pretrained with ImageNet-1K, aswell as Swin-B pretrained on ImageNet-22K. 
We apply the implementation and trained weights from MMSegmentation \cite{contributors2020mmsegmentation}.\\
While our experiments primarily utilize Swin Transformer, we include DINOv2 due to its extensive pretraining and performance on several distortion robustness benchmarks. Only the largest model is trained directly on the dataset. The smaller models utilize knowledge distillation,to facilitate the analysis regarding robust features are retaining during knowledge distillation. %
The trained weights and the implementation provided by the authors are used for our experiments.
\\
As the results on DINOv2 show that iBOT pretraining is beneficial for semantic segmentation, a Swin-T model pretrained with iBOT is used in the training experiment. The trained weights are provided by the authors of iBOT.

%% file: sec/figure_distortion_method.tex
\begin{figure}[t]
    \centering
    %
    \begin{subfigure}{.32\linewidth}
    \includegraphics[width=\linewidth]{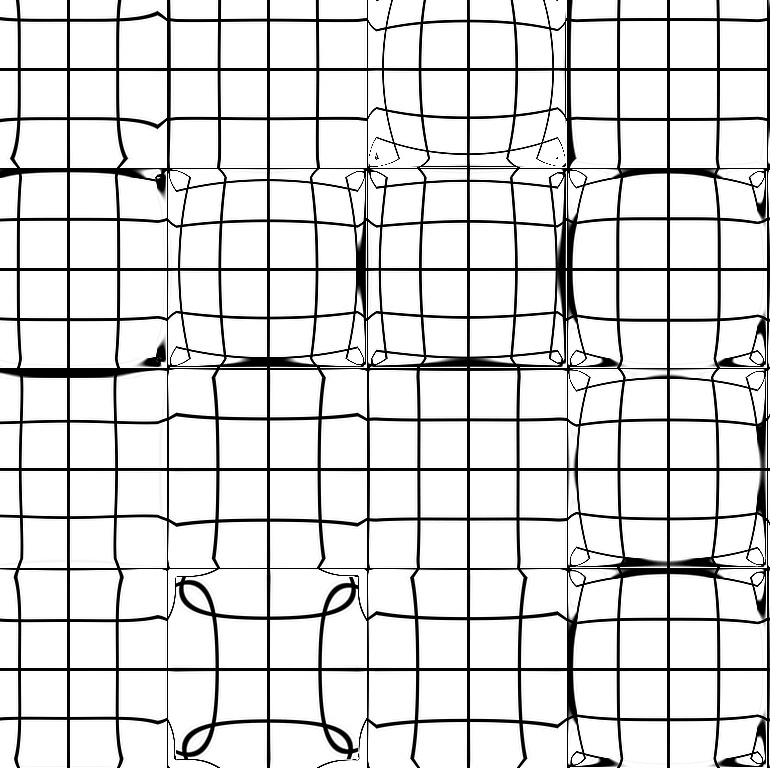}
    \caption{Distorted grid}
    \end{subfigure}
    %
    \hfill
    %
    %
    %
    \begin{subfigure}{.32\linewidth}
    \evalcropcrop{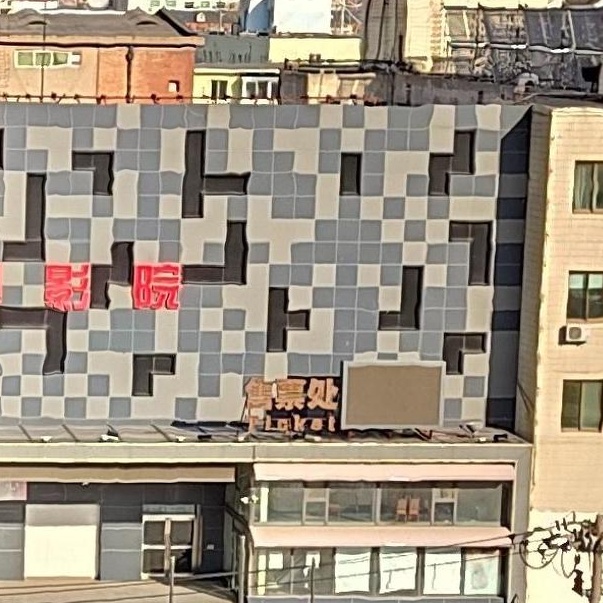}{-2.2cm}{1.9cm}{1.8cm}{1.5cm}{south east}{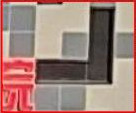}
    \caption{Simulated distortion}
    \end{subfigure}
    \hfill
    \begin{subfigure}{.32\linewidth}    \evalcrop{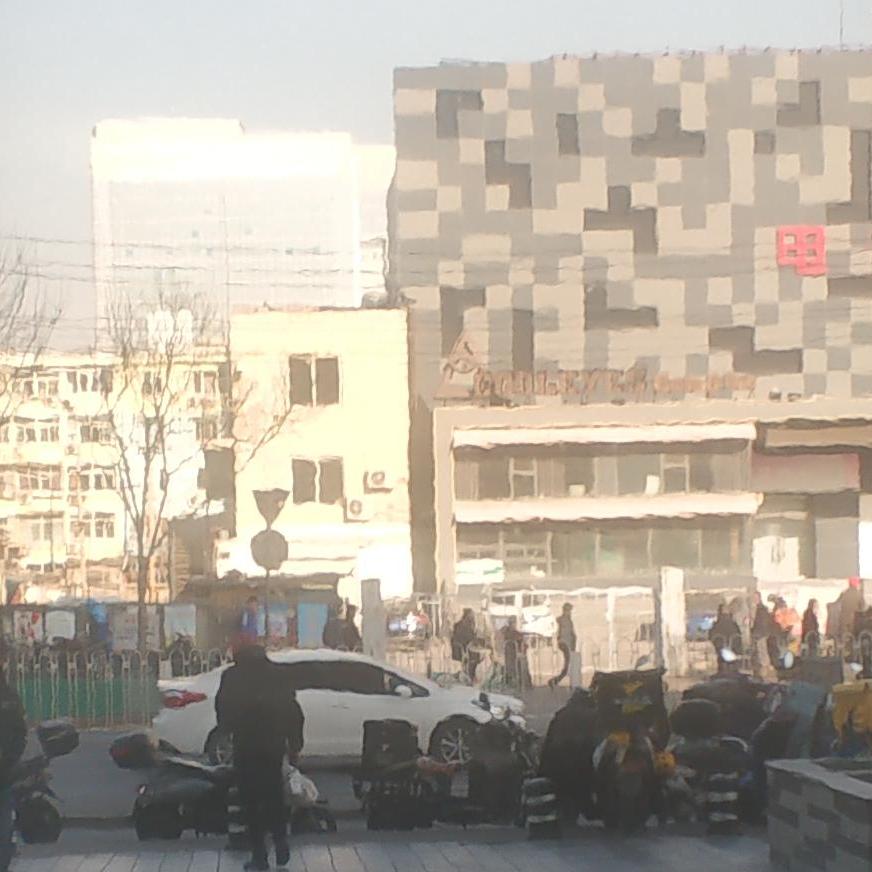}{-0cm}{2.7cm}{1.8cm}{1.5cm}{south east}
    \caption{Real distorted view}
    \end{subfigure}
    \caption{The novel grid distortion model applied to a grid of lines (a) and a real scene (b). 
    The bottom left crop in (b) shows the corresponding undistorted view.
    The simulated distortion of the real scene \cite{xie2020segmenting} creates similarly curved edges as the real distorted view of the same building (c). 
    The real transparent structure also adds other distortions and artifacts like reflection and a lower contrast to the scene, which are not part of the simulation.}
    \label{fig:grid_distortion}
\end{figure}

%% file: sec/figure_global_distortion.tex
\begin{figure}[t]
    \centering
    \begin{subfigure}{.03\linewidth}
        \raisebox{1.8cm}{\rotatebox[]{90}{Global Dist.}}
    \end{subfigure}
    \begin{subfigure}{.31\linewidth}
        \includegraphics[width=\linewidth]{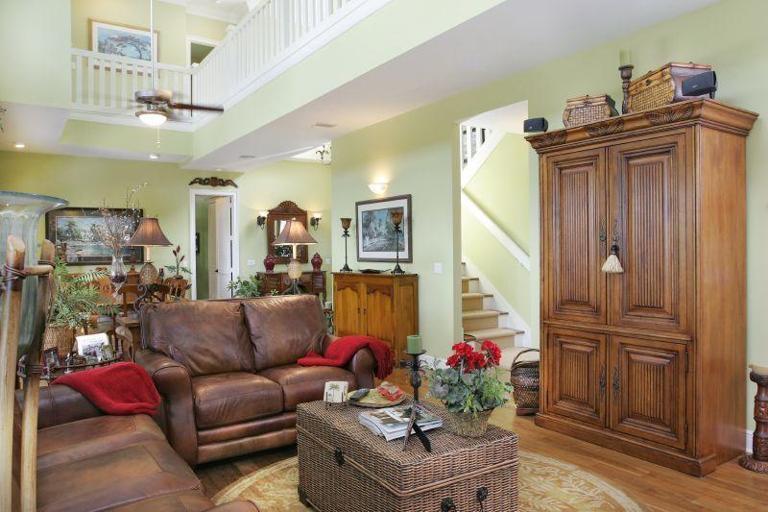}
        \caption{$K_1=K_2=0$ (Orig)}
    \end{subfigure}
    \hfill
    \begin{subfigure}{.31\linewidth}
        \includegraphics[width=\linewidth]{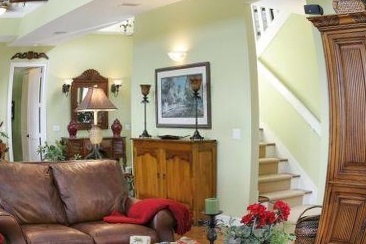}
        \caption{$K_1=K_2=0.4$}
    \end{subfigure}
    \hfill
    \begin{subfigure}{.31\linewidth}
        \includegraphics[width=\linewidth]{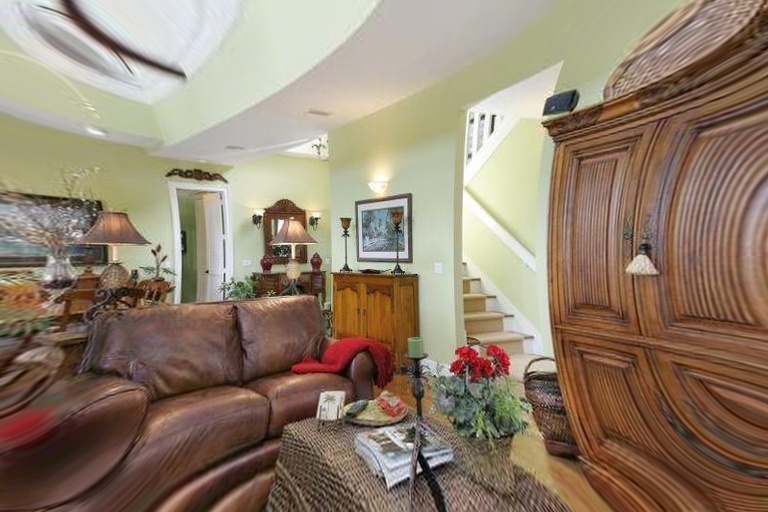}
        \caption{$K_1=-0.25\ K_2=0$}
    \end{subfigure}
    \begin{subfigure}{.03\linewidth}
        \raisebox{1.8cm}{\rotatebox[]{90}{Grid Dist.}}
    \end{subfigure}
    \begin{subfigure}{.31\linewidth}
        \includegraphics[width=\linewidth]{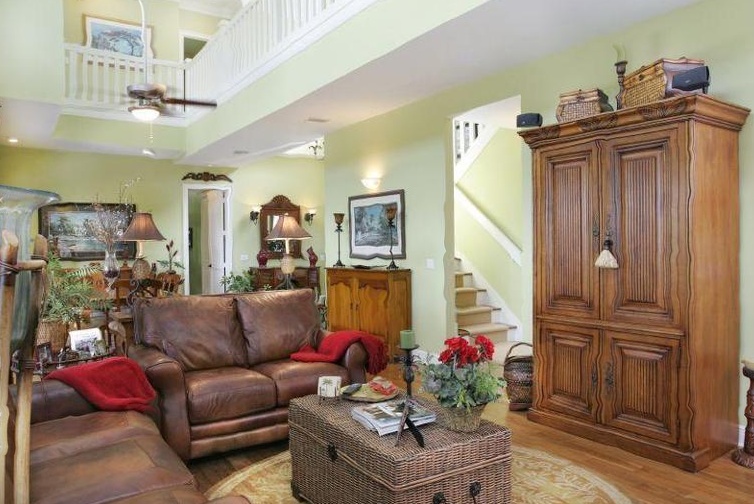}
        \caption{$K_1,K_2 \in [-0.1, 0.1]$}
    \end{subfigure}
    \hfill
    \begin{subfigure}{.31\linewidth}
        \includegraphics[width=\linewidth]{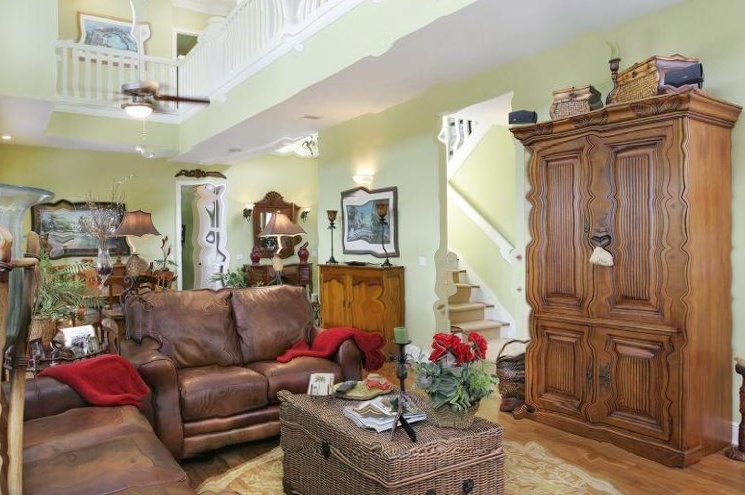}
        \caption{$K_1,K_2 \in [-0.3, 0.3]$}
    \end{subfigure}
    \hfill
    \begin{subfigure}{.31\linewidth}
        \includegraphics[width=\linewidth]{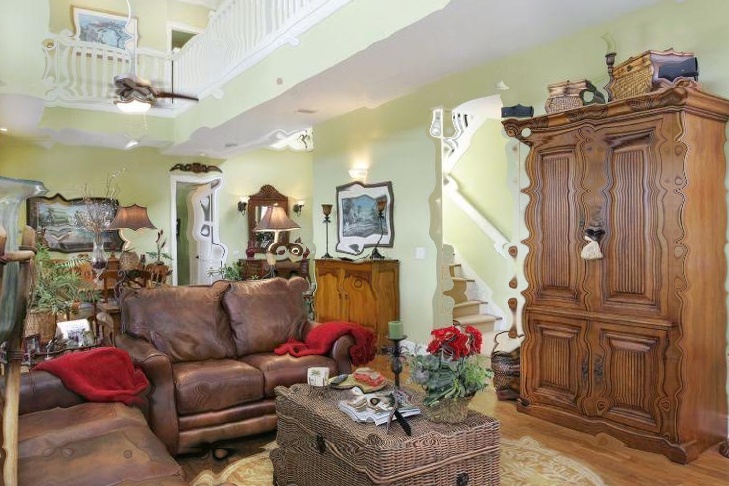}
        \caption{$K_1,K_2 \in [-0.5, 0.5]$}
    \end{subfigure}
    \caption{
    Results of \textit{global radial distortion} (top) and \textit{grid distortion} (bottom) for different parameters $K_1, K_1$ on an example image (a). Global radial distortion with positive parameters (b) leads to barrel distortion, while $K_1 < 0$ and $K_2 = 0$ (c) lead to a severe pincushion distortion effect. With grid distortion increasing values of $K_1, K_2$ lead to increasingly wavy edges (bottom, left to right).
    }
    \label{fig:both_dists}
\end{figure}

%% file: sec/results.tex
\section{Results}

In this section we examine how state-of-the-art semantic segmentation models, originally not tailored or trained for robustness, handle distortions from transparent structures. 
The models are evaluated on the ADE20K validation set, subject to the previously introduced distortion methods. 
This mimics a scenario where a standard model is trained on a conventional dataset featuring clean images and is subsequently confronted with distortions arising from transparent structures in practical applications. 
Both images and the corresponding segmentation masks are distorted identically to ensure alignment of the object boundaries in the distorted images. 

\subsection{Quantitative Results}
\input{sec/iou_plot}
The models are evaluated on the distorted dataset using the mean Intersection over Union (mIoU) metric. 
It describes how well the predicted class masks match the with the ground truth masks in an image.
The IoU is averaged over all classes to give the mean Intersection over Union. 
As this experiment focuses on robustness, the absolute difference in mIoU is compared to the result on the original images. 
This shows how differently each model performs on distorted images, which is a better representation of model robustness. 
\\
Furthermore, the number of different classes predicted in an image is observed. 
This provides insights into the nature of erroneous classifications, as mIoU does not distinguish well between false object boundaries and false class assignments.\\
Figure \ref{fig:iou} illustrates the performance degradation from global (a) and grid (b) distortions. 
All examined models exhibit inferior performance on each distorted dataset compared to the original images. 
Notably, in the context of global distortion, Swin-B demonstrates marginally greater robustness across higher distortion intensities. 
Although larger model sizes seem to contribute to enhanced robustness, the pretraining of Swin-B on the more extensive ImageNet-22K dataset does not manifest a significant impact on performance in this context.
\\
Analyzing the impact of grid distortion (see Fig. \ref{fig:iou}b), it becomes evident that it exerts a more substantial effect on robustness than global distortion.
The difference between the two Swin Transformer sizes is negligible for most distortion intensities.
Notably, the larger pretraining dataset does make a discernible difference in the case of grid distortion, unlike the results observed with global distortion.
Swin-B, pre-trained with ImageNet-22K, displays a less pronounced decline in mIoU across all distortion intensities.
\\
For the DINOv2 models, the decrease in IoU is significantly lower, indicating a substantially higher robustness. 
The huge pre-training dataset used for DINOv2 seems to also result in more robust features for transparent structures. 
Besides that, the knowledge distillation process seems to preserve this robustness as the DINOv2 Base architecture shows similarly good results as DINOv2-G.\\
To provide insights on whether distortions lead to misclassifications or erroneous object boundaries, Table \ref{tab:num-classes} lists the average number of detected classes per image.
No discernible trend is evident, with Swin-T seemingly detecting more classes as distortion intensity increases, while Swin-B detects fewer.
The difference in the number of detected classes in comparison to no distortion is, however, negligible.
The maximum change in the number of detected classes is $7.6\%$ for Swin-T with $\sigma=0.5$.
This suggests that there is robustness in terms of identifying the correct classes, but challenges arise in accurately detecting object boundaries.
Additional details on the segmentation results are given in the following section.

\begin{table}[t]
\centering
\caption[Number of different classes predicted per image on grid distortion]{
    Number of different classes predicted per image on different levels of grid distortion. The ground-truth contains 9.45 classes per image.
}
\setlength{\tabcolsep}{6pt} 
\begin{tabular}{c|c|ccccc}
\toprule
$\sigma$ & 0 & 0.1 & 0.2 & 0.3 & 0.4 & 0.5 \\ \midrule
Swin-T & 10.97 & 11.14 & 11.23 & 11.42 & 11.61 & 11.81 \\
Swin-B & 10.35 & 10.46 & 10.33 & 10.33 & 10.32 & 10.17 \\
Swin-B (22K) & 10.45 & 10.56 & 10.56 & 10.55 & 10.53 & 10.42 \\ \bottomrule
\end{tabular}
\label{tab:num-classes}
\end{table}

\subsection{Qualitative Results} \label{sec:qual}
\input{sec/figure_qualeval}
For the qualitative analysis we show three examples in Fig. \ref{fig:qualeval}, that highlight possible errors that the Swin Transformer exhibited on the distorted ADE20K dataset. 
In comparison to the ground truth of Fig. \ref{fig:qualeval}a, both segmentations encompass the relevant objects. 
However, the segmentation on the distorted image misclassifies a truck as a car. 
This would not pose a major issue until a car predicts a "safe" crash zone through heat shimmers to avoid human casualties. 
Another similar effect can be seen in Fig. \ref{fig:qualeval}b. 
Furthermore, objects are not segmented anymore, additional wrong objects are found and object contours disconnect. 
The last example showcases a bar table in Fig \ref{fig:qualeval}c. 
In this case a large portion of the table is not recognized and a variety of wrong segmentations and classifications are present. 
To summarize our findings, we report four error types that are present between the undistorted and distorted segmentations.
\begin{enumerate}
 \item An object is missing in the segmentation map
 \item An object is detected as a different class
 \item The shape of an object is not correctly detected
 \item The shape of an object is correctly detected but split into multiple classes
\end{enumerate}

%% file: sec/iou_plot.tex
\pgfplotsset{compat=1.11,
    /pgfplots/ybar legend/.style={
    /pgfplots/legend image code/.code={%
       \draw[##1,/tikz/.cd,yshift=-0.25em]
        (0cm,0cm) rectangle (3pt,0.8em);},
   },
}

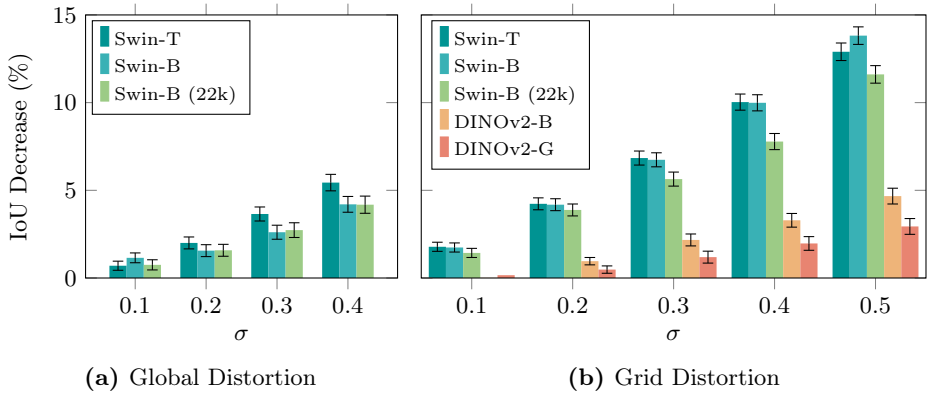
\begin{figure}[t]
    \centering
    \begin{subfigure}{.03\linewidth}
    \raisebox{3.2cm}{\rotatebox[]{90}{IoU Decrease (\%)}}
    \end{subfigure}
    \begin{subfigure}{.34\linewidth}
        \begin{tikzpicture}
        \begin{axis}[mybarplot,
            ymin=0,
            ymax=15,
            xmin=0.03,
            xmax=0.47,
            height=3.5cm,
            xtick={0.1, 0.2, 0.3, 0.4},
            xlabel={$\sigma$},
            bar width=6.5pt,
            legend style={
            at={(0.02,0.98)},
            anchor=north west,
            legend cell align={left},
            font=\scriptsize},
            ]
                \addplot coordinates { (0.1,0.70)+-(0.26, 0.26) (0.2,2.00)+-(0.34, 0.34) (0.3, 3.65)+-(0.4, 0.4) (0.4, 5.44)+-(0.47, 0.47) }; \addlegendentry{1}
                \addplot coordinates { (0.1,1.15)+-(0.28, 0.28) (0.2,1.56)+-(0.34, 0.34) (0.3, 2.61)+-(0.4, 0.4) (0.4, 4.2)+-(0.45, 0.45) }; \addlegendentry{2}
                \addplot coordinates { (0.1,0.75)+-(0.29, 0.29) (0.2,1.58)+-(0.34, 0.34) (0.3, 2.73)+-(0.42, 0.42) (0.4, 4.18)+-(0.49, 0.49) }; \addlegendentry{3}
                \legend{Swin-T, Swin-B, Swin-B (22k)}
        \end{axis}
        \end{tikzpicture}
        \caption{Global Distortion}
    \end{subfigure}%
    \hfill
    \begin{subfigure}{.55\linewidth}
        \hspace{-.3cm}
        \begin{tikzpicture}
        \begin{axis}[mybarplot,
            ymin=0,
            ymax=15,
            xmin=0.05,
            xmax=0.55,
            height=3.5cm,
            xtick={0.1, 0.2, 0.3, 0.4, 0.5},
            xlabel={$\sigma$},
            bar width=6.5pt,
            legend style={
            at={(0.02,0.98)},
            anchor=north west,
            legend cell align={left},
            font=\scriptsize},
            yticklabels={},
            ]
                \addplot coordinates { (0.1,1.78)+-(0.26, 0.26) (0.2,4.23)+-(0.34, 0.34) (0.3, 6.84)+-(0.4, 0.4) (0.4, 10.03)+-(0.46, 0.46) (0.5, 12.9)+-(0.5, 0.5) };
                \addplot coordinates { (0.1,1.74)+-(0.26, 0.26) (0.2,4.18)+-(0.34, 0.34) (0.3, 6.74)+-(0.4, 0.4) (0.4, 9.99)+-(0.46, 0.46) (0.5, 13.82)+-(0.5, 0.5) };
                \addplot coordinates { (0.1,1.43)+-(0.26, 0.26) (0.2,3.88)+-(0.34, 0.34) (0.3, 5.64)+-(0.4, 0.4) (0.4, 7.78)+-(0.46, 0.46) (0.5, 11.61)+-(0.5, 0.5) };
                \addplot coordinates { (0.1,0.01) (0.2,0.96)+-(0.21, 0.21) (0.3, 2.17)+-(0.34, 0.34) (0.4, 3.29)+-(0.39, 0.39) (0.5, 4.67)+-(0.45, 0.45) };
                \addplot coordinates { (0.1,0.16) (0.2,0.48)+-(0.21, 0.21) (0.3, 1.19)+-(0.34, 0.34) (0.4, 1.97)+-(0.39, 0.39) (0.5, 2.94)+-(0.45, 0.45) };
                \legend{Swin-T, Swin-B, Swin-B (22k), DINOv2-B, DINOv2-G}
        \end{axis}
        \end{tikzpicture}
        \caption{Grid Distortion}
    \end{subfigure}%
    \caption{Comparison between Swin Transformer models on different intensities $\sigma$ of \textit{global distortion} (a), and between Swin Transformer models and \textit{DINOv2} on \textit{grid distortion} (b). \textit{Swin-T} and \textit{Swin-B} are the Tiny and Base Swin Transformer models pretrained on ImageNet-1K, \textit{Swin-B (22K)} is the model pretrained on ImageNet-22K. Error bars show the 95\% confidence interval of IoU differences.}
    \label{fig:iou}
\end{figure}

%% file: sec/figure_qualeval.tex
\begin{figure}[t]
\begin{subfigure}{\linewidth}    
    \begin{subfigure}{.03\linewidth}
    \raisebox{1.3cm}{\rotatebox[]{90}{(a) Cars on a Road}}
    \end{subfigure}
    \begin{subfigure}{.31\linewidth}
    \caption*{Ground Truth}
    \includegraphics[width=\linewidth]{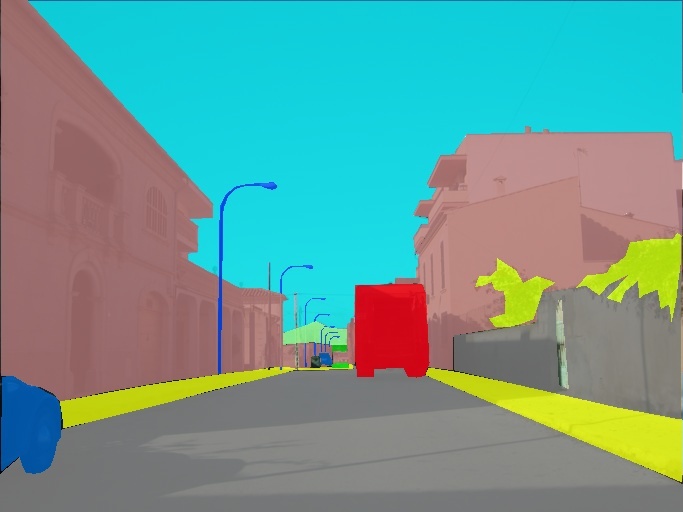}
    \end{subfigure}
    \hfill
    \centering
    \begin{subfigure}{.31\linewidth}
    \caption*{Segmentation}
    \includegraphics[width=\linewidth]{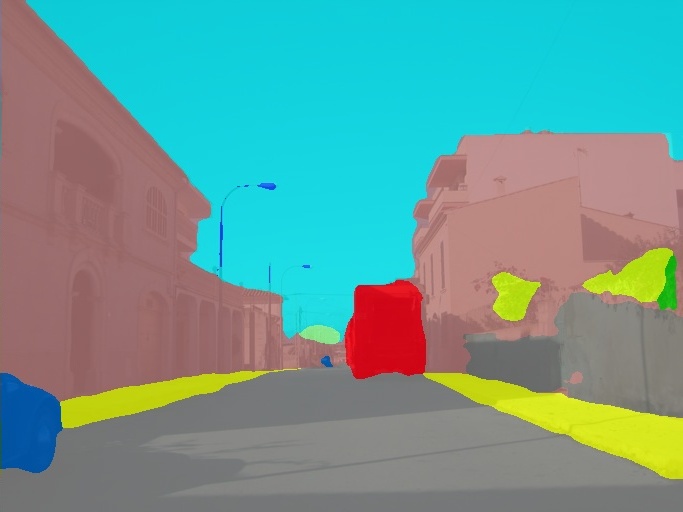}
    \end{subfigure}
    \hfill
    \begin{subfigure}{.31\linewidth}
    \caption*{Distorted Segmentation}
    \includegraphics[width=\linewidth]{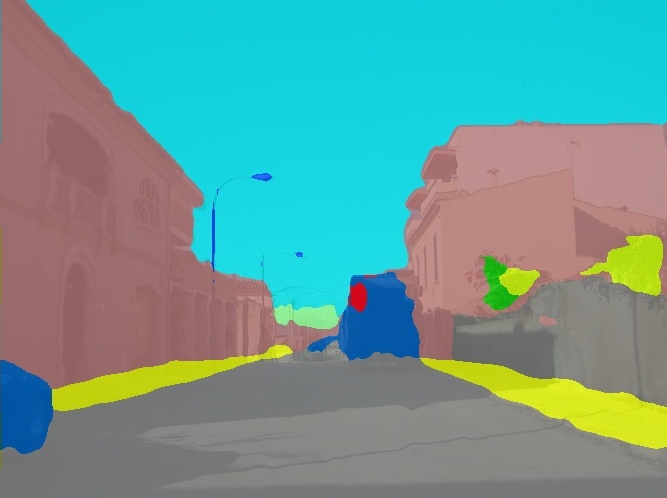}
    \end{subfigure}
\label{Fig:Cars}
\end{subfigure}
    %
    %
\begin{subfigure}{\linewidth}
    \begin{subfigure}{.03\linewidth}
    \raisebox{1.2cm}{\rotatebox[]{90}{(b) Sauna}}
    \end{subfigure}
    \begin{subfigure}{.31\linewidth}
    \includegraphics[width=\linewidth]{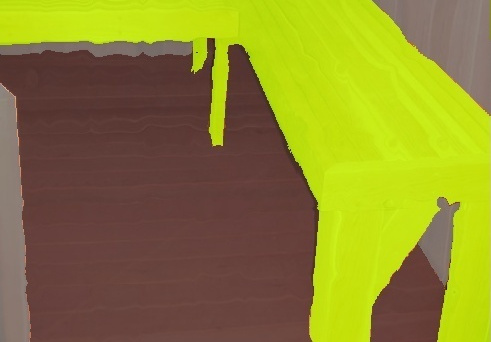}
    \end{subfigure}
    \hfill
    \centering
    \begin{subfigure}{.31\linewidth}
    \includegraphics[width=\linewidth]{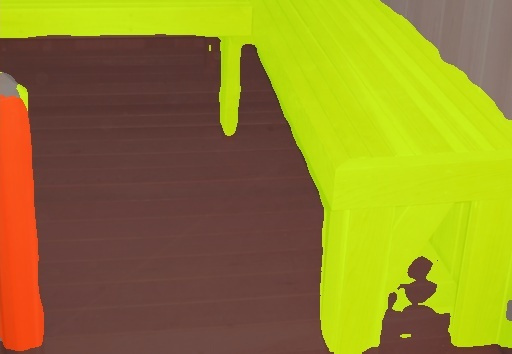}
    \end{subfigure}
    \hfill
    \begin{subfigure}{.31\linewidth}
    \includegraphics[width=\linewidth]{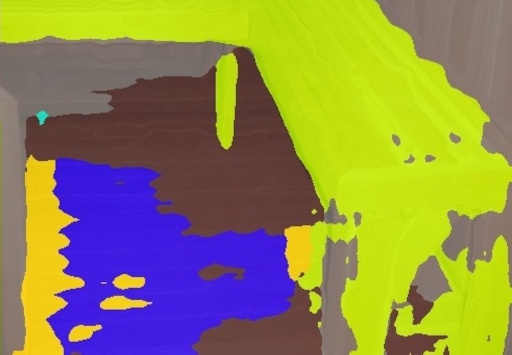}
    \end{subfigure}
\label{Fig:sauna}
\end{subfigure}
    %
    %
\begin{subfigure}{\linewidth}
    \begin{subfigure}{.03\linewidth}
    \raisebox{1.4cm}{\rotatebox[]{90}{(c) Bar Table}}
    \end{subfigure}
    \begin{subfigure}{.31\linewidth}
    \includegraphics[width=\linewidth]{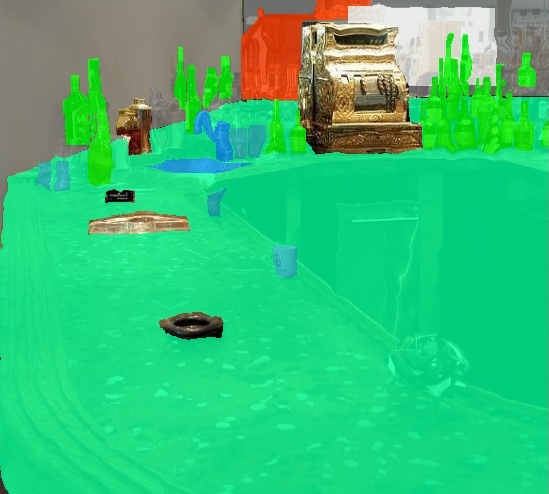}
    \end{subfigure}
    \hfill
    \centering
    \begin{subfigure}{.31\linewidth}
    \includegraphics[width=\linewidth]{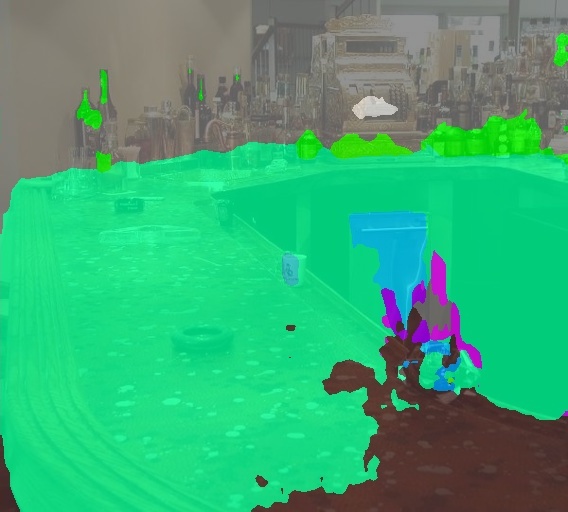}
    \end{subfigure}
    \hfill
    \begin{subfigure}{.31\linewidth}
    \includegraphics[width=\linewidth]{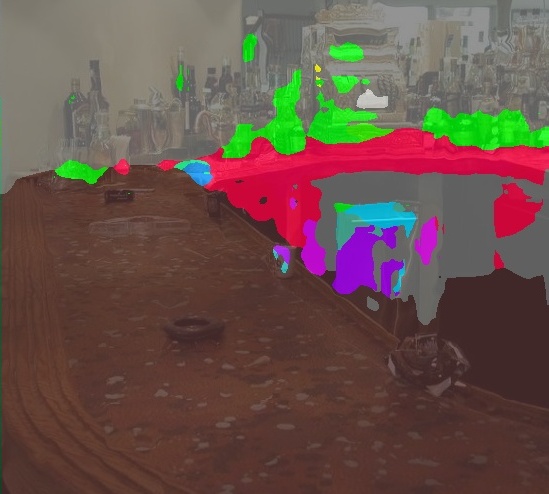}
    \end{subfigure}
\label{Fig:Bar}
\end{subfigure}
    \caption{The ground truth and segmentations on the undistorted and distorted ADE20K datasets \cite{zhou2019semantic}. }
    \label{fig:qualeval}
\end{figure}

%% file: sec/distortiontraining.tex
\section{Distortion Training}
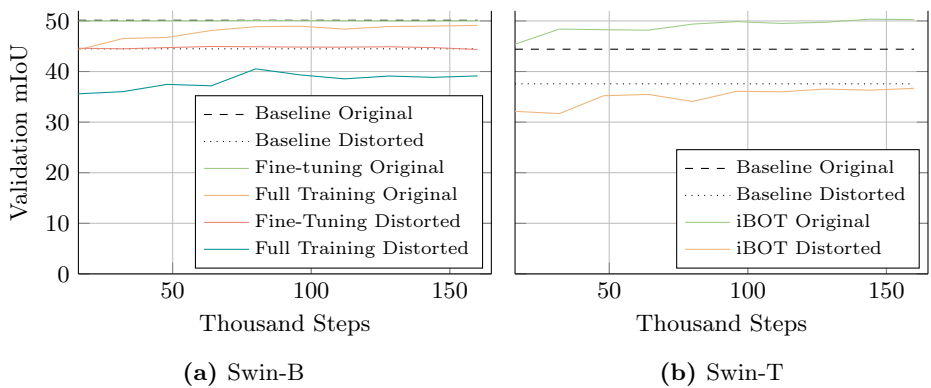
\begin{figure}[t]
    \centering
    \begin{subfigure}{.02\linewidth}
    \raisebox{3.3cm}{\rotatebox[]{90}{Validation mIoU}}
    \end{subfigure}
    \begin{subfigure}{.45\linewidth}
        \begin{tikzpicture}
        \begin{axis}[
            ymin=0,
            ymax=52,
            grid,
            width=\linewidth,
            height=3.5cm,
            scale only axis,
            xmin=16,
            xmax=165,
            ytick={0, 10, 20, 30, 40, 50},
            xlabel={Thousand Steps},
            axis x line*=bottom, 
            legend style={
            at={(0.98,0.02)},
            anchor=south east,
            legend cell align={left},
            font=\scriptsize},
            cycle list={colorscheme0, colorscheme1, colorscheme2, colorscheme4, colorscheme5},
            ]
                \addplot+[mark={}, black, dashed] coordinates { (16, 50.13) (160, 50.13) };
                \addplot+[mark={}, black, dotted] coordinates { (16, 44.49) (160, 44.49) };
                \addplot+[mark={}] coordinates { (16, 50.02) (32, 50.01) (48, 50.02) (64, 50.01) (80, 50.09) (96, 50.05) (112, 50.04) (128, 50.01) (144, 50.00) (160, 50.04) };
                \addplot+[mark={}] coordinates { (16, 44.29) (32, 46.52) (48, 46.74) (64, 48.12) (80, 48.86) (96, 48.96) (112, 48.39) (128, 48.89) (144, 48.98) (160, 49.12) };
                \addplot+[mark={}] coordinates { (16, 44.61) (32, 44.49) (48, 44.73) (64, 44.95) (80, 44.89) (96, 44.83) (112, 44.82) (128, 44.89) (144, 44.73) (160, 44.36) };
                \addplot+[mark={}] coordinates { (16, 35.6) (32, 36.02) (48, 37.46) (64, 37.18) (80, 40.53) (96, 39.33) (112, 38.55) (128, 39.11) (144, 38.85) (160, 39.13) };
                \legend{Baseline Original, Baseline Distorted, Fine-tuning Original, Full Training Original, Fine-Tuning Distorted, Full Training Distorted }
        \end{axis}
        \end{tikzpicture} 
        \caption{Swin-B}
        \label{fig:retb}
    \end{subfigure}%
    \hfill
    \begin{subfigure}{.45\linewidth}
        \hspace{-.3cm}
        \begin{tikzpicture}
        \begin{axis}[
            ymin=0,
            ymax=52,
            grid,
            width=\linewidth,
            height=3.5cm,
            scale only axis,
            xmin=16,
            xmax=165,
            ytick={0, 10, 20, 30, 40, 50},
            xlabel={Thousand Steps},
            axis x line*=bottom, 
            legend style={
            at={(0.98,0.02)},
            anchor=south east,
            legend cell align={left},
            font=\scriptsize},
            yticklabels={},
            cycle list={colorscheme0, colorscheme1, colorscheme2, colorscheme4, colorscheme5},
            ]
                \addplot+[mark={}, black, dashed] coordinates { (16, 44.41) (160, 44.41) };
                \addplot+[mark={}, black, dotted] coordinates { (16, 37.57) (160, 37.57) };
                \addplot+[mark={}] coordinates { (16, 45.39) (32, 48.39) (48, 48.27) (64, 48.17) (80, 49.37) (96, 49.86) (112, 49.51) (128, 49.73) (144, 50.35) (160, 50.26) };
                \addplot+[mark={}] coordinates { (16, 32.13) (32, 31.69) (48, 35.21) (64, 35.48) (80, 34.08) (96, 36.09) (112, 35.99) (128, 36.54) (144, 36.31) (160, 36.68) };
                \legend{Baseline Original, Baseline Distorted, iBOT Original, iBOT Distorted}
        \end{axis}
        \end{tikzpicture} 
        \caption{Swin-T}
        \label{fig:rett}
    \end{subfigure}%
    \caption{Comparison of various retraining methods on original and distorted data.}
    \label{fig:retraining}
\end{figure}

To further emphasize the effect of spatially varying distortion, we report the mIoU metric for several retrained Swin models on the original and distorted ADE20K datasets. 
Fig. \ref{fig:retb} visualizes two graphs for each Swin-B training method depending on the dataset. 
The dotted lines denote the baseline result from training on the original dataset and evaluating on both datasets. 
The other models are either fine-tuned or fully trained on the two datasets.
The results show that neither of those models is able to reach the original baseline. 
Only fine tuning the original on the distorted dataset marginally outperformed the distortion baseline, while fine-tuning the segmentation model on the distorted dataset yields a marginally better performance than the baseline. 
Similar to this approach we trained Swin-T with iBOT pre-training on both datasets, which can be seen in Fig. \ref{fig:rett}. 
The dotted line show the baseline results on the original and distorted dataset for Swin-T with standard ImageNet-1K pre-training. 
While this leads to a performance boost for the original images, it falls short on the distortion baseline. 
In total, our findings emphasize that segmentation issues due to distortion can not be simply solved by retraining on distorted data. 
This leads to the hypothesis, that the original shape of objects is necessary for a NN to understand a segmentation task and that distorted objects are not as easily compared to their original counterparts as they are for humans.

%% file: sec/conclusion.tex
\section{Conclusion and Future Work}
Transparent structures are commonplace in modern society and can cause distortions to the scenes behind them. To simulate this behaviour we introduced a novel grid distortion method am assessed the effect of two distortion types on state-of-the-art semantic segmentation models. We report an overall mIoU performance drop in comparison to the original dataset and that various significant semantic segmentation error categories arise. Additionally, our experiments for robustness improvement training show, that only fine-tuning leads to marginal performance improvements. While larger model sizes and extensive pre-training lead to a better robustness, these improvements depend on the present types of distortions. In conclusion, distortions from common phenomenons like heat shimmers and transparent structures pose a challenge for semantic segmentation tasks. The greater context of our insights and more techniques on distortion mitigation will be developed and evaluated more thoroughly in future work.